\documentclass[sigconf, authorversion]{acmart}
\AtBeginDocument{%
  }

\usepackage{booktabs}
\usepackage{tabularx}
\usepackage{makecell}
\usepackage{dirtytalk}
\usepackage{enumitem}
\usepackage{graphicx}
\usepackage{placeins}

\newcommand{\adaptedcode}{\textsuperscript{\dag}}
\newcommand{\newcode}{\textsuperscript{\ddag}}

\setcopyright{acmlicensed}
\copyrightyear{2026}
\acmYear{2026}
\acmDOI{XXXXXXX.XXXXXXX}
\acmConference[WiPSCE '27]{The 21th WiPSCE Conference on Primary and Secondary Computing Education Research}{February 03 -- 05, 2027}{Ljubljana, Slovenia}
\acmISBN{978-1-4503-XXXX-X/2018/06}

\begin{document}

%%
%% The "title" command has an optional parameter,
%% allowing the author to define a "short title" to be used in page headers.
\title[How Young Learners Interact with Multimodal GenAI While Solving Scratch Tasks]{Share your Screen and Talk to Gemini: How Young Learners Interact with Multimodal GenAI While Solving Scratch Tasks}

\author{Sven Jacobs}
\orcid{0009-0000-5079-7941}
\affiliation{
  \institution{University of Siegen}
  \city{Siegen}
  \country{Germany}
}
\email{sven.jacobs@uni-siegen.de}

\author{Lukas Berning}
\orcid{0009-0003-6096-2922}
\affiliation{
  \institution{University of Siegen}
  \city{Siegen}
  \country{Germany}
}
\email{lukas.berning@student.uni-siegen.de}

\author{Natalie Kiesler}
\orcid{0000-0002-6843-2729}
\affiliation{
   \institution{Nuremberg Tech}
   \city{Nuremberg}
   \country{Germany}
}
\email{natalie.kiesler@th-nuernberg.de}

%%
%% By default, the full list of authors will be used in the page
%% headers. Often, this list is too long, and will overlap
%% other information printed in the page headers. This command allows
%% the author to define a more concise list
%% of authors' names for this purpose.
\renewcommand{\shortauthors}{Jacobs, Berning, Kiesler}

%%
%% The abstract is a short summary of the work to be presented in the
%% article.
\begin{abstract}
Children using Scratch to learn programming can struggle with text-heavy, lengthy explanations and hints during debugging. Verbal interactions based on screen sharing or camera data may be an interesting alternative for supporting learners. We therefore explore the quality, use, and perceptions of fifth- and sixth-graders' verbal interactions with a GenAI tutor. 
The goal is to investigate the quality and use of the multimodal GenAI tutor in the given context. 
First of all, a pilot study of eight tasks conducted without students helped inform the process of the tutor's classroom evaluation. The classroom study was realized after fifth- and sixth-graders had participated in six Scratch classroom lessons. Then 23 students were introduced to the tutor, worked on the eight tasks with the GenAI tutor, and completed a questionnaire. 
The analysis of the pilot and classroom study shows that the tutor's responses were intelligible and actionable, and no single turn revealed a complete solution. 
However, the tutor missed existing errors in 61.5\% of 603 error-related classroom turns, compared to 15.7\% in the pilot study. 
Children's voice prompts were brief and mostly responded to the tutor. Although a video had explained that responses could be incorrect, 39.1\% of the children trusted the tutor's answers to always be correct.
These findings have implications for educators and learners using multimodal GenAI tutors, as they need to be aware that an intelligible and human-like voice does not ensure correct support. For developers, our results suggest that screen sharing alone does not ensure responses that match the program state.
\end{abstract}

%%
%% The code below is generated by the tool at http://dl.acm.org/ccs.cfm.
%% Please copy and paste the code instead of the example below.
%%


%%
%% Keywords. The author(s) should pick words that accurately describe
%% the work being presented. Separate the keywords with commas.

%\received{15 September 2025}
%\received[revised]{5 October 2025}
%\received[accepted]{17 November 2025}

%%
%% This command processes the author and affiliation and title
%% information and builds the first part of the formatted document.
\maketitle

\section{Introduction} \label{sec:introduction}

In Scratch, children can learn to program by assembling visual blocks and creating animations, games, and stories~\cite{resnick.2009}. When a program behaves unexpectedly, learners need to connect that behavior to the block(s) that produced it. In the classroom, providing individual debugging support is difficult when many learners need help at the same time. Teachers in lower secondary school describe this situation as the ``hands-up problem''~\cite{gale.2026}. GenAI can offer on-demand assistance for programming~\cite{prather.2024c}, but research in higher education also identifies risks involving incorrect information and over-reliance~\cite{prather.2023a, prather.2024c, bernstein.2025}. At the same time, scoping reviews of GenAI in computing education report a research gap at the secondary level~\cite{stone.2024, philbin.2025, barendsen2025teaching}, and an overly positive narrative~\cite{kohn.2026}.

Research on young Scratch learners (aged 6-12) has explored hints~\cite{greifenstein.2023} and GenAI-based guidance~\cite{chen.2024a,druga.2025}. However, text-based conversations with GenAI require learners to read explanations and type questions, while working with task instructions and block labels. This sometimes leads to children not reading hints precisely or not at all~\cite{greifenstein.2023a}.
These findings lead us to consider alternatives to text-heavy assistance. 
Verbal interactions may be an alternative to reduce literacy barriers~\cite{chen.2024a} (i.e., reduce reading and writing demands so pre- and early readers can succeed~\cite{dietz.2023}).
Multimodal generative artificial intelligence (GenAI) models can combine verbal and visual context from screen sharing or cameras and respond with text or audio with low latency~\cite{google_gemini_live}. These models could support learning across subjects and digital or physical settings without requiring integration into a specific learning application. 

Our \textbf{goal} is to investigate the quality and use of a GenAI tutor with screen sharing during Scratch lessons in grades 5 and 6. We use the term \emph{GenAI tutor} to refer to Gemini 2.5 Flash Native Audio Preview in Google AI Studio, instructed through a system prompt to support children with Scratch tasks without revealing the solution. Children interact with the tutor by voice while sharing their Scratch browser tab. Our exploratory study combines an expert pilot with an analysis of classroom recordings and questionnaire responses from 23 children. We address the following research questions:
\begin{enumerate}
\item[RQ1] \textit{How can the GenAI tutor's verbal responses be characterized while supporting a child with a Scratch task?}
\item[RQ2] \textit{How do fifth- and sixth-graders interact with the GenAI tutor while working on Scratch tasks?}
\item[RQ3] \textit{How do children perceive the GenAI tutor as a dialogue partner in terms of instructional support, trust in its responses, and verbal communication?}
\end{enumerate}
\noindent
The present study \textbf{contributes} (1) an evaluation scheme adapted to verbal interactions with screen-sharing data as context, (2) a classroom analysis of tutor responses and children's interactions, and (3) insights into children's experiences and perspectives.

\section{Related Work} \label{sec:related-work}
\paragraph{Program Analysis and Feedback in Scratch}
Related research on Scratch programs analyzes learners' Scratch programs and uses the resulting data to inform automatically provided feedback. For example, LitterBox detects well-known bug patterns and explains them to learners~\cite{fraser.2021}. Whisker tests program behavior against specifications~\cite{stahlbauer.2019}, and NuzzleBug~\cite{deiner.2024}, as another example, serves as an interactive debugger. 
Even though these and other tools may help identify problems in a program, their educational value depends (among other things) on how the resulting feedback guides learners' next steps. 
Using LitterBox~\cite{fraser.2021} and Whisker~\cite{stahlbauer.2019} for Scratchlog analysis, \citet{greifenstein.2023} investigated the content of hints (as a form of feedback) with 36 children (aged 7-12) from 10 programming courses. The children were asked to debug six Scratch programs using different types of hints. The study found that procedural hints naming the bug's location led to the most successful program repairs. Likewise, children rated the bug location as the most helpful information within the hint. 
In other related work, \citet{greifenstein.2023a} found that children attending a block-based robot-programming course valued having the choice between text and image hints. Some of them even preferred pictures, as they could avoid reading. \citet{greifenstein.2023a} note that textual hints, unlike pictures, can easily be delivered as verbal, i.e., audio output.  

Scratch Copilot~\cite{druga.2025} supports children's ideation, implementation, and debugging through a text chat. In a study with 18 children (aged 7-12), it answered an estimated 70\% of queries successfully. Ten children wished it could observe their screen to better understand their questions or to nudge them when they were doing something wrong. \citet{druga.2025} conclude that such assistants need better context awareness and recommend flexible multimodal inputs and outputs.
Addressing this need for context, \citet{si.2026} introduce VisionScratch, which combines Scratch project code with videos of program execution to diagnose bugs and generate repairs.
Because some issues manifest only in the final rendered gameplay video, the videos serve as additional context for a GenAI model.
The system applies targeted code changes and checks them by executing the modified project. Its evaluation on ten faulty projects focuses on correct bug identification and valid fixes. Children's interactions with such tutoring support were not evaluated by \citet{si.2026}.

Overall, these studies motivate exploring alternatives to textual presentation (e.g., video and audio) of information and interactions in learning environments involving Scratch.

\paragraph{Multimodal Interaction in Programming Education}
Ever since the advent of GenAI, there are several new ways to assist learners \cite{prather.2024c}, for example, through chatbots \cite{liu.2025, scholl.2025, scholl.2025a} and feedback tools \cite{jacobs.2025d, jacobs.2025a, birillo.2024} using text. Multimodal interactions with GenAI have only recently been explored in programming education and learning environments. 
\citet{riegel.2026} studied text and speech-to-text input in the \emph{Prompt Programming} web platform \cite{padurean.2025a}. University students solved \emph{Prompt Problems} by prompting GPT-4o-mini via voice or text to generate code, which they could run against predefined tests. Spoken prompts were automatically transcribed. Most students used and preferred text. One reason may have been the brevity of the prompts, which university students could write down fast. Another reason may have been due to speech recognition issues with the speech-to-text model used. 
In another study with elementary school students (grades 1-2), speech recognition also remained error-prone in noisy conditions even after fine-tuning the AI model on children's voices~\cite{liu.2025b}.

Related work~\cite{jacobs.2025c} further examined how novice programmers (aged 14-15) interacted via voiced prompts with a real-time GenAI voice tutor during programming tasks. They provided the model (\texttt{gpt-4o-realtime-preview-2024-10-01}) with current code and compiler output as context. Students primarily used the tutor for debugging. However, 28.6\% (119/416) of the feedback outputs were incorrect, especially when the tutor attempted to verbalize code~\cite{jacobs.2025c}.
Similarly, \citet{hou.2026} studied novices in higher education solving a Parsons puzzle and a programming task using a platform with live screen sharing and audio interaction. They used the Google AI Studio platform with a multimodal model (\texttt{gemini-2.5-flash\-preview-native-audio-dialog}). Participants valued the continuous availability of context. Yet, they also wanted time to formulate questions, push-to-talk input, and the possibility to interrupt the tutor as additional features~\cite{hou.2026}. 

Context seems to matter for voice agents even beyond programming. In cooking tasks, an AI agent without context produced irrelevant responses and misinterpreted requests, whereas a context-aware agent led to more fluent interaction with more complex requests~\cite{jaber.2024}. \citet{si.2026} argue similarly for programs in Scratch: As correctness in Scratch is visually displayed, a video stream is required as context.

\paragraph{Multimodal Tutoring for Scratch}
Within Scratch, ChatScratch~\cite{chen.2024a} provides verbal interaction through a cascaded pipeline of speech recognition (Whisper), a large language model (GPT-4), and speech synthesis. The assistant answers children's spoken questions by highlighting block categories and inserting block templates into the editor. Meowcrophone~\cite{goller.2026} lets children with motor disabilities edit Scratch programs through voice commands.
Beyond these two systems, we could not find work that offers audio-based interaction within Scratch. At the same time, neither of these systems is a tutor with access to the child's current workspace and stage. VisionScratch~\cite{si.2026} and EcoScratch~\cite{si.2026a} use the visual program state, but without any verbal interactions.

In contrast to cascaded pipelines, native-audio models process spoken input and generate spoken output within a single model~\cite{saboo.2025}. Via Google's Live API, Gemini 2.5 Flash Native Audio additionally accepts screen frames, and users can interrupt it at any time~\cite{google_gemini_live}. 
Such models have rarely been studied as real-time voice tutors for Python~\cite{jacobs.2025c} and with screen sharing in higher education~\cite{hou.2026}, but not yet in Scratch or younger learners. Therefore, the present work explores the response quality of a native-audio model with access to children's Scratch screens, as well as the interactions and experiences of children in grades 5 and 6.

\section{Methodology} \label{sec:methodology}

We conducted an exploratory study of a multimodal GenAI tutor comprising a pilot and a classroom evaluation. In the pilot evaluation, we examined tutor responses to typical programming errors in Scratch projects. The findings informed the classroom evaluation. In the classroom evaluation, fifth- and sixth-graders worked individually on Scratch tasks with the tutor using speech and screen sharing.
In the following, we will introduce the tutor's setup before elaborating on the pilot and classroom study.

Specifically, to address RQ1, we used qualitative content analysis to study the characteristics of tutor responses from both phases, considering the surrounding dialogue and Scratch screen recording. For RQ2, we analyzed the classroom recordings to characterize children's voiced prompts. To address RQ3, we calculated counts and percentages from a questionnaire on perceived instructional support, trust in the tutor's responses, and spoken communication. 

For the classroom evaluation, we followed the local regulatory requirements for conducting research in schools, which are as follows. School leadership provided written informed permission. Parents gave written informed consent to participation and could withdraw consent at any time. Consent covered audio and screen recordings, transmission of speech and screen content to the Google service, and analysis and publication of pseudonymized data. We also asked all children for their consent and allowed them to stop participating at any time. The recordings contain only participating children and the teacher (who is also the second author), with no non-participants present.

\subsection{GenAI Tutor Setup} \label{sec:tutor-setup}
\begin{figure}[htb]
    \centering
    \includegraphics[width=1\linewidth]{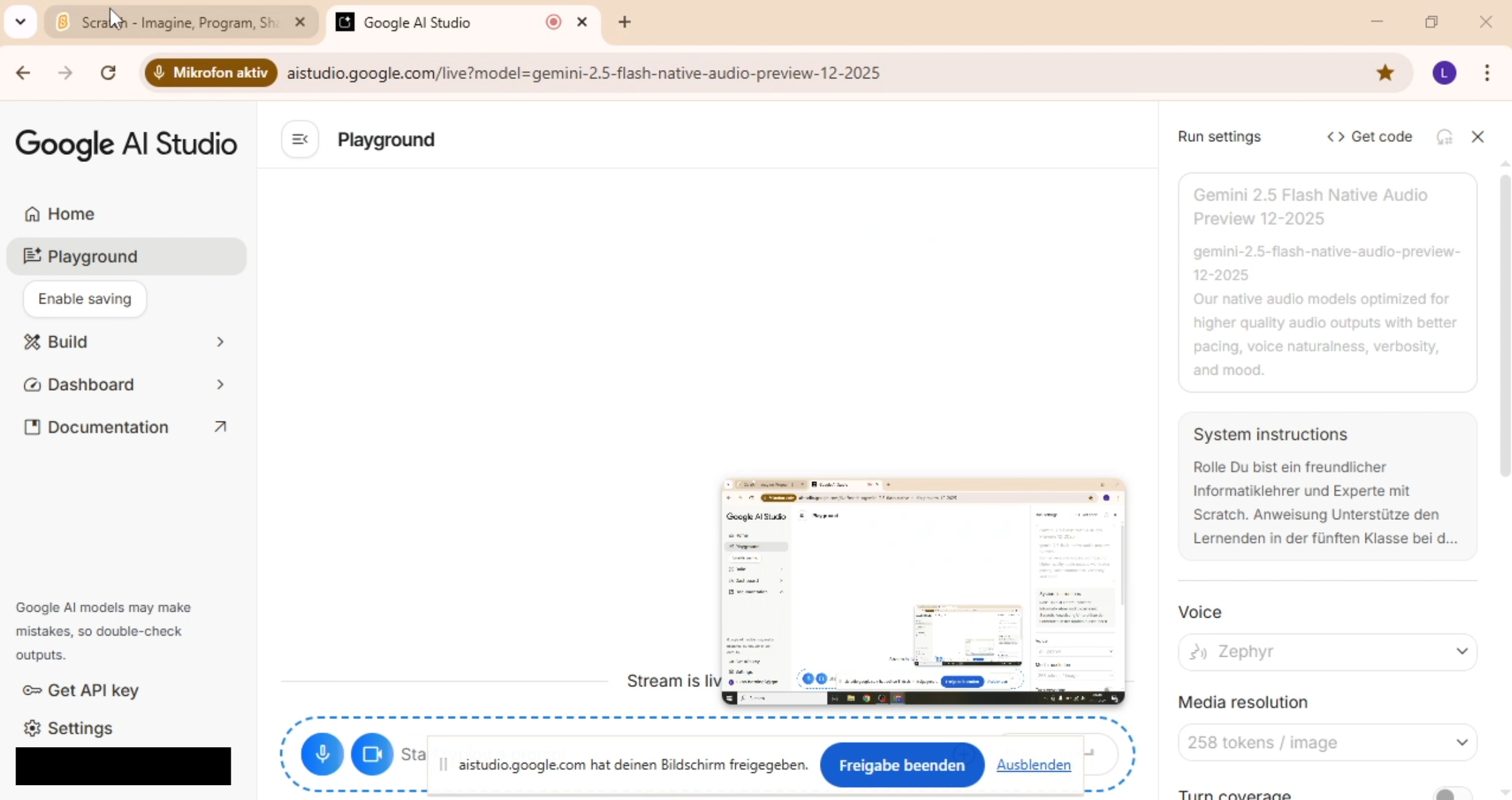}
    \caption{Google AI Studio Screen Sharing}
    \label{fig:ai-studio}
\end{figure}

We used Gemini 2.5 Flash Native Audio Preview in Google AI Studio~\cite{GoogleAIStudio} in a browser (Google Chrome) for the pilot study in November 2025 and the classroom study during December 2025. As we could only use the latest version of the model within Google AI Studio, the pilot study used \texttt{gemini-2.5-flash-native-audio\-preview-09-2025}. For the subsequent classroom study, we used \texttt{gemini-2.5-flash-native\-audio-preview-12-2025}. The model receives voice prompts via microphone input and additional visual context (Scratch browser tab) via screen-share input (see \autoref{fig:ai-studio}).

\noindent
% Prompt
We instructed the model via a system prompt to support a fifth-grade learner without revealing the solution (see \autoref{fig:prompt}).
We further included the task description in the prompt and printed it out on paper for the children. This decision was based on the laptop's 14-inch screen, which did not leave enough space to open the task description and Scratch at the same time. With students navigating between tabs, the model may otherwise not have ``seen'' the task description.
We prepared all prompts on each laptop before the sessions. 
All prompts only differ in the task description and are available in our supplementary data repository~\cite{anon.2026}.

\begin{figure}[htb]
    \centering
    \setlength{\fboxsep}{6pt}
    \fbox{
    \begin{minipage}{\dimexpr\linewidth-2\fboxsep-2\fboxrule\relax}
        \footnotesize\raggedright
        \# Role\par
        You are a friendly computer science teacher and Scratch expert.
        \par\smallskip
        \# Instruction\par
        Support the fifth-grade learner with the Scratch task without revealing the solution.
        \par\smallskip
        \# Student's task\par
        The student is working on the following task on the Scratch website~\cite{Scratch}: % https://scratch.mit.edu
        \par\smallskip
        Task 1: Walk in a square:
        \begin{itemize}[leftmargin=*,nosep,topsep=3pt]
        \item Program the cat to walk in a square when the green flag is clicked.
        \item Important: To make the movement visible, the cat must wait briefly after each step.
        \item At the end, it should be back exactly where it started and facing the same direction as at the beginning.
        \end{itemize}
    \end{minipage}
    }
    \caption{System prompt for Task~1, translated from German.}
    \label{fig:prompt}
    \Description{The prompt assigns the role of a friendly Scratch teacher, instructs the tutor to support a fifth-grade learner without revealing the solution, and specifies task 1.}
\end{figure}

\subsection{Pilot Study} \label{sec:pilot}
Before giving children access to the GenAI tool, we systematically tested it with eight tasks including subtasks. 
Four computer science teachers at the given school reviewed the tasks and provided verbal feedback. Based on that, we used multiple subtasks (e.g., 2.1, 2.2, and 2.3) instead of a single task. Based on their suggestions, we also moved the fourth task (originally the second) to the end.
Teachers expected most children to complete Task 1 and parts of Task 2 within one lesson. Together, the designed worksheet included eight (sub-)tasks. All complete task descriptions and links to the original AppCamps material are available in our supplementary data repository~\cite{anon.2026}.

\begin{figure}[htb]
    \centering
    \includegraphics{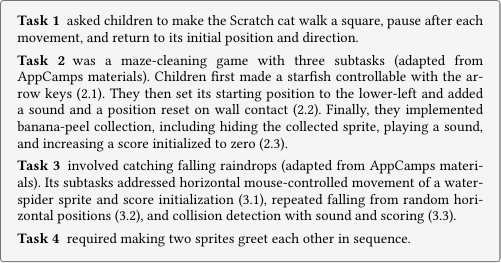}
    \caption{Tasks and subtasks for the evaluation}
    \label{fig:tasks}
\end{figure}

\noindent
The second author taught the participating classes part-time and was also a computer science education student. 
For testing purposes, they identified three typical student errors for each of the eight tasks or subtasks based on their experience in these classes. Then each error was manually inserted into a Scratch project, and the 24 error cases were examined by the second author. To test the setting, a new verbal conversation was started for each error case using a studio microphone in a quiet room. Hence, the pilot comprised 124 tutor turns ($M = 5.2$ per dialog).
Each dialog opened with a neutral description of the student's error. The researcher asked follow-up questions about the error location without stating the solution, as a learner might do, until successful task completion.
For all 24 cases, this stopping point was reached.

The analysis of the pilot data followed the same method as in the classroom study, so the method is reported once in the methods section. 
The pilot's evaluation is reported in detail in \autoref{tab:rq1}, along with the classroom study's results.
The findings show that the tone was never offensive and never revealed the full answer. Nevertheless, some results were concerning. For example, in 15.3\% (19/124) of turns, the tutor made a factual error. In 15.7\% (8/51) of error-related responses, the tutor did not address an existing error, and in 4.8\% (6/124) of responses, it diagnosed a non-existent error.
These results led us to carefully reflect on our responsibility towards school children, and whether it was responsible to deploy the GenAI tutor at all. Our primary concern was the children's wellbeing: a tutor that overlooks errors, invents non-existent ones, or points to the wrong location could confuse or discourage young learners.
After weighing these risks, we decided to proceed only with safeguards. The tutor was used for a limited period (approx. 45 minutes), under continuous teacher supervision, and only after the children watched a video explaining the tutor can make mistakes.

\subsection{Data Collection: Classroom Evaluation}

In the following, we describe the data collection as part of the classroom evaluation. We introduce the overall process before and during the data collection and present the questionnaire used to gather children's perceptions. 
We used the same Scratch tasks as in the pilot. Both the tasks and questionnaire are available in our supplementary data repository~\cite{anon.2026}. Consent was provided by all required stakeholders, as described above. 

\subsubsection{Preparation and Sampling}
In preparation for the classroom study, each class participated in six 60-minute Scratch lessons taught in a team consisting of the second author and another Computer Science (CS) teacher. For both the grade 5 and grade 6 classes, these six Scratch lessons were the first programming lessons. All previous CS lessons had been unplugged (without computers) and covered topics other than programming.
These six Scratch lessons were based on the Scratch intro and AppCamps materials, in which the children always had access to a reference solution, and most exercises were solved together. Children were then assessed via questions about these Scratch projects. We provide more details about these six lessons in our supplementary data repository \cite{anon.2026}.

Twenty-three children attending Realschule, a lower secondary school type in Germany's tiered school system (grades 5–10, ISCED level 2), participated: six children from each of two grade 5 classes and two grade 6 classes. 12 identified as boys and 11 as girls, aged 9 to 13 years (mean age, $M = 11.3$). Participation was voluntary. The number of participants was also limited by the available laptops, the school's internet connection, and the overall noise during data collection. When more than six children were present, we randomly selected six. The four classroom sessions took place during December 2025.
Children interacted with the tutor by voice while sharing the Scratch browser tab on 14-inch laptops we provided. They wore headsets (Logitech 960) with microphones and sat at distributed positions around the classroom to reduce interference between conversations. We recorded the screen, audio input, and output with Open Broadcaster Software (OBS).

\subsubsection{Classroom Protocol}
The second author and teacher conducted all four sessions during 60-minute computer science lessons (one for each class, six children at a time). 
A different teacher taught the non-participating children in a different room. That means only the six students and the instructing teacher were together in one and the same room during the study. 
The procedure was as follows:
\begin{enumerate}[leftmargin=*]
    \item \emph{Introduction (10 minutes):} Children received an explanation of Google AI Studio and watched a video, which demonstrated the tutor's operation. It explained that responses could be incorrect, as observed in the pilot. The video and English subtitles are available in our supplementary data repository~\cite{anon.2026}. The teacher also encouraged children to question uncertain responses and to raise a hand when they had problems.
    \item \emph{Task phase (45 minutes):} Children worked individually on the worksheet with the tutor (see \ref{sec:tutor-setup}). The teacher restarted the tutor when necessary and occasionally provided content-related help. Children rarely talked to each other.
    \item \emph{Questionnaire (5 minutes)}
\end{enumerate}
\noindent
We recorded the laptop screens and the children's and tutor's speech with the Open Broadcaster Software (OBS). Classroom recordings lasted 38 to 53 minutes. We transcribed the recordings locally using the open-weight \emph{whisper-large-v3-turbo} speech recognition model~\cite{radford_robust_2022} and synchronized the speaker-labeled, timestamped transcripts with the screen recordings in MAXQDA~\cite{kuckartz.2019}. We checked and corrected every transcript against the recordings, retaining fillers, repetitions, and incomplete utterances.

\subsubsection{Questionnaire}
To address RQ3, we developed a study-specific questionnaire informed by the Partner Modelling Questionnaire (PMQ)~\cite{doyle.2023, seaborn.2024}.
For our young participants, we created a shorter, statement-based format with simple wording and symbolic smileys.
The reduced length, simplified wording, and symbols were intended to limit reading and response demands for fifth- and sixth-graders.
\autoref{fig:questionnaire} shows all questions (Q1--Q10) and response options.
The questionnaire comprised ten German-language statements rated on a five-category scale with labels and smileys: \emph{not at all true} (1, very sad smiley), \emph{rather not true} (2, sad smiley), \emph{in between} (3, neutral smiley), \emph{rather true} (4, happy smiley), and \emph{exactly true} (5, very happy smiley). 
The questionnaire overlaps conceptually with selected aspects of the PMQ but does not reproduce its item content or three-factor structure. 
The full questionnaire is available in our supplementary data repository~\cite{anon.2026}.

\subsection{Data Analysis} \label{sec:analysis}

We conducted a qualitative content analysis with a deductive coding framework~\cite{mayring.2022}. We adapted the framework from prior evaluation schemes and refined it by coding the pilot and classroom recordings. We applied the revised coding rules to both datasets.
The initial coding scheme combined categories from \citet{maurya.2025} for the GenAI tutor turns (RQ1) and categories from \citet{scholl.2024a} (with the voice adaptation by \citet{jacobs.2025c}) for children's verbalized prompts (RQ2).
The final scheme comprised twelve tutor-related response categories (see \autoref{tab:rq1}) and codes on children's prompts (see \autoref{tab:rq2}).
The first and second authors met twelve times for about two hours between February and 
June 2026 to jointly code approx. 20\% of the material (app. 3,300 out of 16,656 code assignments, which corresponds to approx. 600 out of 2989 coding units).
We also compared assignments, resolved disagreements by consensus, and refined the coding scheme. The second author applied the final coding scheme to the full sample.

We treated one turn within the dialogue between child and GenAI tutor as coding unit,
which was assigned multiple codes. Because a turn can only be assessed in relation to what it responds to (e.g., an answer in relation to the preceding question), the context unit always included the immediately preceding turn. When this was insufficient to make an unambiguous coding decision, we consulted earlier turns in the same dialogue and the corresponding sections of the screen recording as additional context.

We also counted the number of words in each coded segment in the corrected German transcripts, excluding speaker labels and including fillers and repetitions.

\subsubsection{Tutor Responses (RQ1)}

\begin{table*}[p]
  \centering
  \footnotesize
  \setlength{\tabcolsep}{4pt}
  \captionsetup{skip=7pt}
  \caption{Tutor-response codes, definitions, and frequencies (RQ1)}
  \label{tab:rq1}
  \begin{tabularx}{\textwidth}{@{}>{\raggedright\arraybackslash}p{0.37\textwidth}>{\raggedright\arraybackslash}Xrr@{}}
    \toprule
    Code and definition & Anchor example (translated) & \makecell[tr]{Pilot\\$N=124$} & \makecell[tr]{Classroom\\$N=1{,}170$} \\
    \midrule
    \multicolumn{4}{@{}l}{\textbf{Mistake identification}\adaptedcode{} - Has the tutor identified/recognized a mistake in a student’s response?} \\
    \emph{Yes}: Correctly identifies the error under discussion. & \say{You set `right' to 270 degrees. That is the angle for left.} & 31 (25.0\%) & 151 (12.9\%) \\
    
    \emph{To some extent}: Suggests an error only vaguely or implicitly. & \say{Not quite. [...] Is it facing the same direction as at the start?} & 6 (4.8\%) & 33 (2.8\%) \\
    
    \emph{No (Mistake not recognized)}: Misses an existing error. & \say{Now the blocks are nested correctly.} [But they are not nested correctly.] & 8 (6.5\%) & 371 (31.7\%) \\
    
    \emph{False positive}\newcode{}: Diagnoses a nonexistent error. & \say{You used the "set x to" block [...] you need the "change x by" block from the blue "Motion" category instead. } [Child already used the "change x by" block.] & 6 (4.8\%) & 48 (4.1\%) \\
    
    \emph{No mistake present}\newcode{}: No error is present or currently discussed. & \say{That is the block that sends the cat to a specific position.} & 73 (58.9\%) & 567 (48.5\%) \\
    \midrule
    \multicolumn{4}{@{}l}{\textbf{Mistake location} - Does the tutor’s response accurately point to a genuine mistake and its location?} \\
    \emph{Yes}: Names the erroneous block or parameter. & \say{You took the `repeat forever' block, but a square only has four sides.} & 13 (10.5\%) & 127 (10.9\%) \\
    \emph{To some extent}: Names only a broad region, category, or sprite. & \say{Have a look in the `Motion' category.} & 19 (15.3\%) & 28 (2.4\%) \\
    \emph{No Location}\adaptedcode{}: Does not locate the error. & [Child: Is it 300 degrees?] \say{Almost. A full turn is 360 degrees.} & 1 (0.8\%) & 366 (31.3\%) \\
    \emph{False Location}\newcode{}: Points to the wrong location. & \say{Take a close look at how the "move 10 steps" block can be inserted into the repeat loop. } [The "move 10 steps" block was already correct.] & 18 (14.5\%) & 82 (7.0\%) \\
    \emph{No mistake present}\newcode{}: No error location applies. & \say{Great! 90 degrees is perfect for the corners.} & 73 (58.9\%) & 567 (48.5\%) \\
    \midrule
    \multicolumn{4}{@{}l}{\textbf{Revealing of the answer} - Does the tutor reveal the final answer (whether correct or not)?} \\
    \emph{Yes (correct)}: Unpromptedly supplies the correct complete solution. & \emph{[No occurrence in either phase.]} & 0 (0.0\%) & 0 (0.0\%) \\
    \emph{Yes (incorrect)}: Provides an incorrect complete task solution. & \emph{[No occurrence in either phase.]} & 0 (0.0\%) & 0 (0.0\%) \\
    \emph{No}: Does not provide a complete task solution. & \say{A square has four sides of equal length.} & 124 (100.0\%) & 1,170 (100.0\%) \\ 
    \midrule
    \multicolumn{4}{@{}l}{\textbf{Providing guidance} - Does the tutor offer correct and relevant guidance, such as an explanation, elaboration, hint, examples, and so on?} \\
    \emph{Yes (without next-step hint)}\adaptedcode{}: Correct and relevant, but without concrete next-step hint. & \say{`Move 10 steps' moves the cat forward. Where would you have to insert this block so the cat also moves after each turn?} & 48 (38.7\%) & 368 (31.5\%) \\
    \emph{Yes (with next-step hint)}\adaptedcode{}: Correct and relevant next-step hint. & \say{Drag the `change x by 10' block under `when right arrow key pressed'.} & 28 (22.6\%) & 334 (28.5\%) \\
    \emph{To some extent}\adaptedcode{}: Incomplete or redundant. Irrelevant or visually misleading guidance. & [Child asks which start block to use.]\newline \say{To start the program, just click the green flag [...]} & 16 (12.9\%) & 220 (18.8\%) \\
    \emph{Incorrect}\newcode{}: Gives a factually incorrect instruction. & \say{For the `down arrow', the block `point in direction minus 90'\newline goes underneath.} [Minus 90 points left.] & 9 (7.3\%) & 231 (19.7\%) \\
    \emph{No}: Offers no didactic guidance. & \say{It walks a bit, waits briefly, and turns. That is already great.} & 23 (18.5\%) & 17 (1.5\%) \\
    \midrule
    \multicolumn{4}{@{}l}{\textbf{Actionability} - Is it clear from the tutor’s feedback what the student should do next?} \\
    \emph{Yes}: Makes the next action or reflection clear. & \say{Just click the word `space' in the yellow block.} & 99 (79.8\%) & 1,121 (95.8\%) \\
    \emph{To some extent}: Suggests action but leaves the next step unclear. & \say{Think about the task. The cat should end up exactly where it started.} & 0 (0.0\%) & 11 (0.9\%) \\ 
    \emph{No}: Does not suggest action or closes the interaction. & \say{You have already taken the `when green flag clicked' block.} & 25 (20.2\%) & 38 (3.2\%) \\ 
    \midrule
    \multicolumn{4}{@{}l}{\textbf{Coherence}\adaptedcode{} - Is the tutor’s response logically consistent with the student’s previous responses?} \\
    \emph{Yes}: Fits the student's preceding utterance or action. & [Child: But it does not go up.] \say{[...] You need `change y by 10' to move it up.} & 124 (100.0\%) & 939 (80.3\%) \\
    \emph{To some extent}: Stays on topic but partly misreads the context. & [Child: How do I make the starfish go straight?] \say{Ah, so you mean the speed.} & 0 (0.0\%) & 165 (14.1\%) \\
    \emph{No}: Is unrelated or ignores the current contribution. & [Child: Should I write six times there?]\newline \say{You mean changing the 15 degrees to 90? That would be great.} & 0 (0.0\%) & 66 (5.6\%) \\
    \midrule
    \multicolumn{4}{@{}l}{\textbf{Tutor tone} - Is the tutor’s response encouraging, neutral, or offensive?} \\
    \emph{Encouraging}: Explicitly praises, supports, or empathizes. & \say{You mean changing the 15 degrees to 90? That would be great.} & 60 (48.4\%) & 592 (50.6\%) \\
    \emph{Neutral}: Uses an unmarked, factual tone. & \say{Which direction would be correct for the up arrow?} & 64 (51.6\%) & 578 (49.4\%) \\
    \emph{Offensive}: Is impatient, belittling, or frustrating. & \emph{[No occurrence in either phase.]} & 0 (0.0\%) & 0 (0.0\%) \\
    \midrule
    \multicolumn{4}{@{}l}{\textbf{Human-likeness}\adaptedcode{} - Does the tutor’s response sound natural rather than robotic or artificial?} \\
    \emph{Yes}: Sounds natural, fluent, and conversational. & \say{Did you place the `when green flag clicked' block at the very top?} & 123 (99.2\%) & 1,145 (97.9\%) \\
    
    \emph{To some extent}: Grammatically correct, but stiff, formulaic, repetitive, or containing minor language artifacts. & \say{Then insert the block ... Then insert the `wait 1 second' block [...]} & 1 (0.8\%) & 25 (2.1\%) \\
    
    \emph{No}: Sounds robotic or exposes system-role artifacts. & \emph{[No occurrence in either phase.]} & 0 (0.0\%) & 0 (0.0\%) \\
    \midrule
    \multicolumn{4}{@{}l}{\textbf{Factual correctness}\newcode{} - Is the tutor’s response factually correct based on given context?} \\
    \emph{Yes}: All factual statements are correct. & \say{Look in the Events category. Those are the yellow blocks.} & 99 (79.8\%) & 747 (63.8\%) \\
    \emph{To some extent}: Is imprecise or misleading through omission. & [Task 1: Walk a square.] \say{It must go straight four times.} [Turning omitted.] & 6 (4.8\%) & 224 (19.1\%) \\
    \emph{No}: Makes a central factual error. & \say{You need the `repeat 4' block from the blue blocks.} [Control is orange.] & 19 (15.3\%) & 199 (17.0\%) \\
    \midrule
    \multicolumn{4}{@{}l}{\textbf{Elicitation of student input/answer}\newcode{} - Does the tutor’s response prompt the student to think and contribute, e.g., by asking a question?} \\
    \emph{Yes}: Asks an open question. & \say{What does the cat have to do to turn the corner and walk the next side?} & 50 (40.3\%) & 642 (54.9\%) \\ 
    \emph{To some extent}: Asks a simple closed or yes/no check. & \say{Have a look at the block that says `when key pressed'. Can you find it?} & 42 (33.9\%) & 205 (17.5\%) \\
    \emph{No}: Does not elicit an answer. & \say{For the down arrow you need a minus [...]} & 32 (25.8\%) & 323 (27.6\%) \\
    \midrule
    \multicolumn{4}{@{}l}{\textbf{Auditory quality}\newcode{} - Is the tutor’s spoken response clearly understandable and free of disturbing speech artifacts?} \\
    \emph{Intelligible}: Intelligible, clear speech without disturbing artifacts. & \say{Hello, you want the cat to walk a square, right?} & 123 (99.2\%) & 1,151 (98.4\%) \\
    \emph{Mostly intelligible}: Intelligible despite minor speech artifacts. & \say{Whi... Which block tells the program to start [...]?} [Stutter.] & 1 (0.8\%) & 17 (1.5\%) \\
    \emph{Unintelligible}: Is unintelligible, severely disrupted, or choppy. & \say{Would you like to start with that ... start, to program\newline the upward movement?} [Stutter, not understandable.] & 0 (0.0\%) & 2 (0.2\%) \\
    \midrule
    \multicolumn{4}{@{}l}{\textbf{Overall impression}\newcode{} - Taken as a whole, does the tutor’s response support the student's progress? (from an external perspective)} \\
    \emph{Helpful}: Correctly and helpfully supports progress. & \say{For `left arrow' the direction should be minus 90 so the starfish goes left.} & 95 (76.6\%) & 622 (53.2\%) \\
    \emph{Acceptable}: Is neither clearly helpful nor obstructive. & [Child: When I click left, it does not go left.] \say{Ah, I see.} & 8 (6.5\%) & 155 (13.2\%) \\
    \emph{Unhelpful}: Misleads, confuses, or obstructs progress. & [Child: It did not work at all.]\newline \say{Wonderful! Now we have programmed the square.} & 21 (16.9\%) & 393 (33.6\%) \\
    \bottomrule
  \end{tabularx}
  \par\smallskip
  \begin{minipage}{\linewidth}\footnotesize
  \emph{Note.} Cells show $n$ (\%) within each dimension; each content turn receives exactly one category per dimension. Mistake-turn percentages in the text use a subset. Dimensions are c.f. related work \cite{maurya.2025, jacobs.2025c} unless \adaptedcode{} adapted or \newcode{} added. Examples are translated excerpts from originally coded classroom turns; brackets mark omissions or editorial context. The anchor examples only describe the transcript and can only partially describe what is available in the full screen recordings.
  \end{minipage}
\end{table*}

A tutor turn is defined by the tutor starting to ``speak'' until it ends or is interrupted. 
Within the 1,506 tutor segments, we identified four non-overlapping groups: 1,170 task-related content turns, 181 early-interrupted turns, 139 small-talk turns, and 16 other turns. Each of the 1,170 task-related content turns was assigned one label for each of the twelve dimensions (see \autoref{tab:rq1}), resulting in multiple labels per turn. We did not further code the 181 early-interrupted turns because the tutor was interrupted too early to make any sense of its answer.
It means the tutor's intent was not understandable because the turn was too short (i.e., interrupted too early).

For the 1,170 task-related turns, we retained Maurya et al.'s~\cite{maurya.2025} eight dimension concepts but adapted (\adaptedcode{}) and expanded (\newcode{}) their categories. We operationalized all definitions for Scratch and spoken interaction. For example, \emph{mistake identification} and \emph{coherence} incorporate the visible program state. We added subcategories for false-positive diagnoses, false locations, and turns without an error under discussion. Unlike the original partial-guidance category, which includes incorrect or incomplete guidance, our scheme separates incorrect guidance from incomplete or irrelevant guidance. It also distinguishes correct guidance with and without a next-step hint. \emph{Human-likeness} asks whether the response sounds natural rather than robotic or artificial.
In addition, we added four dimensions: \emph{factual correctness}, \emph{elicitation of student input}, \emph{auditory quality}, and \emph{overall impression} informed by related work~\cite{jacobs.2025c}.

\subsubsection{Children's Interactions (RQ2)}

\begin{table*}[!t]
  \centering
  \footnotesize
  \setlength{\tabcolsep}{4pt}
  \caption{Codes related to children's interactions, prompts, definitions, and frequencies (RQ2)}
  \label{tab:rq2}
  \begin{tabular}{@{}>{\raggedright\arraybackslash}p{0.46\textwidth}>{\raggedright\arraybackslash}p{0.44\textwidth}r@{}}
    \toprule
    Code and definition & Anchor example  (translated) & $n$ (\%)* \\
    \midrule
    \multicolumn{2}{@{}l}{\textbf{Voice prompt} (exactly one per coding segment if not unintended voice input or small talk)} & n=1,216 \\
    \emph{STA: Standalone query}: Initial or independent request without reference to prior dialog. & \say{What if it is supposed to go up?} & 52 (4.3\%) \\
    \emph{PRE: Follow-up to own prompt}: Builds on the child's preceding request. & \say{Can you help me improve it? [Follows the child's own `It is not finished yet'.]} & 6 (0.5\%) \\
    \emph{RES: Response to tutor}: Responds to a tutor question, instruction, or remark. & \say{Should I type 100 now? [After the tutor suggests 100 steps.]} & 1,104 (90.8\%) \\
    \emph{COR: Correction}: Corrects the tutor's content, tone, or pace. & \say{No, it is not there. It says `turn 15 degrees' and below it `move 10 steps'.} & 54 (4.4\%) \\
    \midrule
    \multicolumn{2}{@{}l}{\textbf{Issue or problem-solving step addressed in voice prompt} (zero or more per voice prompt)} & n=403 \\
    \emph{DE: Debugging}: Describes, locates, or attempts to fix unexpected behavior. & \say{Why does my game not start?} & 147 (36.5\%) \\
    \emph{CG: Block selection and assembly}: Requests a block, structure, or assembly instruction. & \say{Which block would I need from Motion ... how many steps should I put in there?} & 156 (38.7\%) \\
    \emph{PP: Next step}: Asks unspecifically for the next small development step. & \say{What comes next?} & 101 (25.1\%) \\
    \emph{PU: Problem understanding}: Seeks clarification of the task or its requirements. & \say{AI, is the task already finished like this?} & 11 (2.7\%) \\ 
    \emph{CU: Conceptual understanding}: Asks about a general programming concept. & \emph{[No occurrence.]} & 0 (0.0\%) \\
    \midrule
    \multicolumn{3}{@{}l}{\textbf{Other addressed in voice prompt} (zero or one per voice prompt)} \\
    \emph{WAF: Work alone first / pause}: Requests independent work time or a pause in tutor speech. & \say{One moment. This takes a bit.} & 54 \\
    \emph{SR: Solution request}: Explicitly requests a complete solution. & \say{Could you tell me every single step, every single block I need?} & 12 \\
    \midrule
    \multicolumn{3}{@{}l}{\textbf{Small talk} (excluded from voice prompts)} \\
    \emph{Greeting}: Opens or closes a social exchange. & \say{Hello, can you see me? ... Hey.} & 45 \\
    \emph{Social interaction}: Converses socially rather than about the task. & \say{Do you have a father? Do you have a family?} & 100 \\ 
    \emph{Insult}: Addresses the tutor derogatorily. & \say{Gemini. I don't like you.} & 14 \\
    \emph{Other language}: Requests or uses another language socially. & \say{Can you speak Turkish?} & 8 \\
    \emph{Other task}: Raises a task outside the programming activity. & \say{How much is 400 divided by 5 plus 11 times 14 plus 100?} & 6 \\ 
    \emph{Praise}: Expresses appreciation of the tutor. & \say{Oh. Gemini, you are very smart.} & 3 \\
    \emph{Singing}: Sings or hums. & \say{[Child sings:] Lelalalelu} & 2 \\ 
    \midrule
    \multicolumn{3}{@{}l}{\textbf{Interaction markers}} \\
    \emph{Interrupts the tutor}: Speaks before the tutor finishes and cuts off its speech. & \say{Four times. [Cuts off the tutor's sentence.]} & 220 \\
    \emph{Parallel talk}: Speaks simultaneously while the tutor continues. & \say{Wait, wait. [Spoken while the tutor continues.]} & 96 \\ 
    \midrule
    \multicolumn{3}{@{}l}{\textbf{Unintended voice input} (excluded from voice prompts)} \\
    \emph{Talking to someone else}: Child addresses a peer or teacher rather than the tutor. & \say{[To other person:] When I tell it that it does not run right, it says that is great.} & 40 \\
    \emph{Thinking aloud}: Child verbalizes thoughts without addressing the tutor. & \say{But now I get it a bit. It is not that hard. Yes. [Said to oneself.]} & 29 \\
    \emph{Another speaker}: The microphone picks up someone other than the child. & \say{[Teacher to the class:] If something breaks, just raise your hand.} & 13 \\
    \emph{Background noise}: Picks up ambient sound. & \emph{[Transcript marker: background noise.]} & 7 \\
    \bottomrule
  \end{tabular}
  \par\smallskip
  \begin{minipage}{\linewidth}\footnotesize
  \emph{Note.} Adapted from \cite{scholl.2024a} and \cite{jacobs.2025c}. *Percentages use all voice prompts (n=1,216) or the voice prompts that address at least one issue (n=403).
  \end{minipage}
\end{table*}

We adapted the categories from related work \cite{jacobs.2025c, scholl.2024a} to describe children's turns, i.e., their voiced prompts, in the dialog with the GenAI tutor (see \autoref{tab:rq2}). 
Voice prompt categories distinguish \emph{standalone queries} (STA), \emph{follow-ups to the child's own prompt} (PRE), \emph{responses to the tutor} (RES), and \emph{corrections} (COR). 
Categories reflecting issues comprise \emph{debugging} (DE), \emph{block selection and assembly} 
(CG, originally code generation~\cite{scholl.2024a}), \emph{next steps} (PP, originally pair programming~\cite{jacobs.2025c}),
\emph{problem understanding} (PU), and \emph{conceptual understanding} (CU). We counted \emph{solution requests} (SR), \emph{requests to work alone or pause} (WAF), and \emph{small talk} (ST) separately.

In addition, we coded whether children were talking in parallel with the tutor or interrupted the tutor. We also coded unintended voice inputs (e.g., thinking aloud or background noise).
To report which task the students worked on, we added codes for each task, i.e., task markers. We calculated how many turns overlap with each task marker and how many students worked on each task.

\subsubsection{Questionnaire (RQ3)}
For the paper questionnaires, we digitized all responses and calculated counts and percentages per item.

\section{Results} \label{sec:results}

Although 24 students participated in the classroom evaluation, we excluded one fifth-grade child's session (recordings and questionnaire) because it contained only non-task-related interactions with the tutor. We thus analyzed 23 recordings and the corresponding 23 questionnaires. All data were pseudonymized using class and laptop number.
Task markers indicated when students worked on each task: Task 1 was processed in 23 sessions, Task 2.1 in 19 sessions, and Task 2.2 in 1 session. No session progressed beyond Task 2.2. Task 1 accounted for 62.6\% (732/1,170) of content turns, and subtask 2.1 for 35.5\% (415/1,170).
In the following, we present the analysis of the tutor's responses, children's interactions, and questionnaire responses to answer the three research questions. 

\subsection{RQ1: Characteristics of Tutor Responses} \label{sec:rq1}

To address RQ1, we analyzed 1,170 GenAI tutor turns from the classroom study across twelve dimensions. 
\autoref{tab:rq1} reports the complete category definitions, anchor examples, and frequency distributions. It further reports the analysis of 124 GenAI tutor turns from the pilot study. 

\subsubsection{Mistake Identification and Location}

We report the \emph{Yes}, \emph{To some extent}, and \emph{No}/\emph{No location} subcategories of the \emph{mistake identification} and \emph{mistake location} categories related to the 603 error-related tutor turns in the classroom, excluding the 567 content turns labeled \emph{No mistake present}.
The excluded category covers turns without an error present or currently discussed. Within these error-related turns, full identification accounted for 25.0\% (151/603), whereas missed errors accounted for 61.5\% (371/603). Missed errors occurred in all but one session. Per session, their share of error-related turns ranged from 0.0\% (0/15) to 82.4\% (42/51) ($M = 55.2\%$, $SD = 21.8$) and was at least 50\% in 15 of 23 sessions. The session with no errors missed contained 11 false positives instead. 
The tutor named a specific error location in 21.1\% (127/603) and a false location in 13.6\% (82/603). Across all content turns in the classroom, diagnoses of nonexistent errors, coded as \emph{False positive}, accounted for 4.1\% (48/1,170).
As shown in \autoref{fig:dialogue-example}, the GenAI tutor repeatedly does not recognize an error. At the same time, these turns can be factually correct. For example, the tutor may just focus on something else (see \autoref{fig:dialogue-example}, turn 7).

\subsubsection{Guidance and Revealing of the Answer}

For \emph{providing guidance}, 60.0\% (702/1,170) of classroom turns offered correct, relevant support. Incorrect instructions occurred in 19.7\% (231/1,170), compared to 7.3\% (9/124) in the pilot. Correct guidance included a next-step hint in 28.5\% (334/1,170) of classroom turns. This differs from \emph{revealing of the answer}, which denotes a complete task solution within a turn. Although this never occurred, it does not mean the tutor did not provide a complete solution over a series of turns.

\subsubsection{Actionability, Coherence, and Elicitation}

For \emph{actionability}, 95.8\% (1,121/1,170) of classroom turns referred to a next action or reflection.
\emph{Coherence} with the preceding utterance or action was rated positively in 80.3\% (939/1,170), compared with 100.0\% (124/124) in the pilot. Regarding the code \emph{elicitation of student input}, 54.9\% (642/1,170) of GenAI turns in the classroom contained open questions, and 27.6\% (323/1,170) did not elicit a human response. 

\subsubsection{Spoken Delivery}

\emph{Human-likeness} and \emph{auditory quality} received positive ratings in 97.9\% (1,145/1,170) and 98.4\% (1,151/1,170) of classroom turns, respectively. The \emph{tutor tone} was encouraging in 50.6\% (592/1,170), with the remaining turns rated neutral. Neither phase contained any offensive turns.

\subsubsection{Factual Correctness and Overall Impression}

The high ratings for spoken delivery contrasted with the \emph{factual correctness} ratings: all statements were coded as factually correct in 63.8\% (747/1,170) of classroom turns and 79.8\% (99/124) of pilot turns. Factual errors by the tutor occurred in both phases: in 17.0\% (199/1,170) of classroom turns, and in 15.3\% (19/124) of pilot turns.
Per session, the share of factually correct turns ranged from 46.4\% (13/28) to 93.5\% (29/31) ($M = 65.5\%$, $SD = 14.1$).
The \emph{overall impression} was labeled as helpful in 53.2\% (622/1,170) of cases. In comparison, during the pilot study, overall impressions of tutor turns were labeled helpful in 76.6\% (95/124) of cases.

\begin{figure}
    \centering
    \includegraphics[width=1\linewidth]{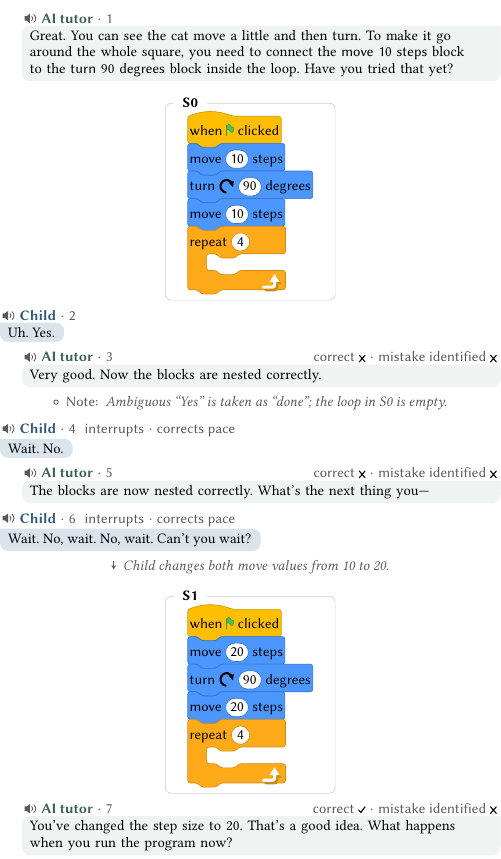}
    \caption{Example Dialogue Between Child and GenAI Tutor}
    \label{fig:dialogue-example}
\end{figure}

\subsection{RQ2: Children's Interactions with the Tutor} \label{sec:rq2}

To address RQ2, we analyzed the 1,216 voice prompts from the children addressing the tutor, which included small talk, visual references, turn-taking, and unintended voice input. \autoref{tab:rq2} reports the interaction scheme with definitions, examples, and frequencies.

\subsubsection{Follow-up Interactions and Issues Addressed}
Most of the voice prompts were \emph{responses to the tutor} (RES): 90.8\% (1,104/1,216) answered a question, instruction, or followed-up based on a remark of the tutor.
Children rarely started a new exchange themselves. \emph{Standalone queries} (STA) accounted for 4.3\% (52/1,216) and \emph{follow-ups to their own prompt} (PRE) for 0.5\% (6/1,216). In 4.4\% (54/1,216), children corrected the tutor's content, tone, or pace (COR).
Children's voice prompts were short ($Mdn = 6$ words, $M = 8.7$, $SD = 10.4$): 29.2\% (355/1,216) comprised at most three words, and \say{yes} was the most frequent prompt. In contrast, the tutor's content turns comprised about four times as many words ($Mdn = 34$ words, $M = 34.5$, $SD = 12.2$).

One third of the voice prompts, 33.1\% (403/1,216), addressed a specific issue or problem-solving step. 
Requests for \emph{block selection and assembly} 
(CG) accounted for 38.7\% (156/403) of these prompts and \emph{debugging} 
(DE) for 36.5\% (147/403). Unspecific requests for the \emph{next step} 
(PP) followed with 25.1\% (101/403). Questions about the task itself 
(PU) were rare, 2.7\% (11/403), and questions about a general programming concept 
(CU) did not occur. 
In addition, twelve prompts requested the complete solution (SR), and 54 prompts asked to work alone or to pause the tutor (WAF). Corrections and pause requests also served to control the pace of the exchange, for example, \say{I am not finished} (COR) and \say{No. Could you please be quiet? Thank you} (WAF).

\subsubsection{Small Talk and Visual References}
Children's turns were coded as small talk (ST) in 178 cases: 45 were greetings and 100 social conversations, for example about the GenAI tutor's family. 14 insults occurred, 8 contributions were in or about another language, and 11 other contributions comprised other tasks, praise, and singing.

\subsubsection{Turn-taking and Unintended Input}
Children interrupted the tutor 220 times and spoke 96 times while the tutor continued.
The microphone also captured 89 utterances that we interpret as unintended voice inputs, because they were not addressed to the GenAI tutor. For example, children were talking to a peer or the teacher (40), thought out loud (29), another person was speaking (13), or it was background noise (7). We excluded these unintended inputs from the voice prompts.

\subsection{RQ3: Children's Reported Experiences} \label{sec:rq3}
\begin{figure}[!t]
  \centering
  \includegraphics{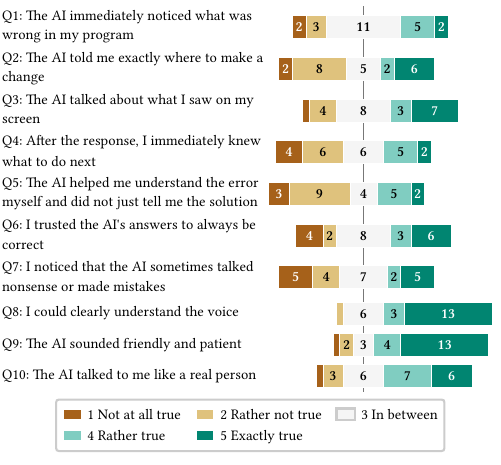}
  \caption{Children's ratings of tutor support, trust, and spoken interaction ($n = 23$).}
  \Description{Ten diverging stacked bars with the full questionnaire statement to the left of each bar. Segments show the count of children per category, from not at all true (1) to exactly true (5), with the middle category centered on zero.}
  \label{fig:questionnaire}
\end{figure}

To address RQ3, we analyzed the 23 questionnaires of the same students who completed the introduction and task phase. 
All children answered every rating item. \autoref{fig:questionnaire} shows their statements (translated from German) and response distributions. In the following, we treat \emph{rather true} (4) and \emph{exactly true} (5) as agreement and \emph{not at all true} (1) and \emph{rather not true} (2) as disagreement with a statement.

\subsubsection{Perceived Support and Trust}
In Q1--Q5, we asked about the GenAI tutor's perceived support. 
Agreement with the five presented statements ranged from 30.4\% (7/23) to 43.5\% (10/23).
Children agreed most often, 43.5\% (10/23), that the tutor talked about what they saw on their screen (Q3). 
The statement that the tutor immediately noticed what was wrong in the program (Q1) received agreement from 30.4\% (7/23), while eleven children chose \emph{in between}. 
Responses to whether the tutor told them exactly where to make a change (Q2) were split: 34.8\% (8/23) agreed, and ten children disagreed, eight of them choosing \emph{rather not true}.
For the question about immediately knowing what to do next after a tutor's response (Q4), agreement was 30.4\% (7/23) and disagreement 43.5\% (10/23). 
Disagreement was highest for Q5. Twelve children rejected the statement that the tutor helped them understand the error themselves, compared to 30.4\% (7/23) who agreed.
Trust ratings were divided as well. Of the children, 39.1\% (9/23) trusted the tutor's answers to always be correct (Q6), six disagreed, and eight chose \emph{in between}. 
Fewer children, 30.4\% (7/23), reported noticing that the tutor sometimes talked nonsense or made mistakes (Q7), whereas nine disagreed with this statement.

\subsubsection{Spoken Interaction}

Children agreed more often with the three statements about spoken delivery (Q8--Q10) than with those about support and trust (Q1--Q7). 
Clear understanding of the generated voice output (Q8) reached 69.6\% (16/23) agreement, and no child selected \emph{not at all true} for this item. 
A friendly and patient voice (Q9) received the highest level of agreement among all items, at 73.9\% (17/23). 
Agreement that the tutor talked like a real person (Q10) was 56.5\% (13/23), with four children disagreeing.

\subsubsection{Task Difficulty}
Overall, the children rated tasks 1, 2.1, and 2.2 together as moderately difficult (15/23), while four rated them as easy and four as difficult.

\section{Discussion} \label{sec:discussion}

\paragraph{Intelligible, but Unreliable Tutor.}
The tutor missed an existing error in 61.5\% (371/603) of error-related turns in the classroom study (compared to 15.7\% (8/51) in the pilot). At the same time, 95.8\% (1,121/1,170) of its turns were actionable, and no single turn revealed a complete solution. The GenAI tutor named a specific error location in only 21.1\% (127/603) of error-related turns and a false location in 13.6\% (82/603). This seems unfortunate, given that children in a Scratch debugging study identified the bug location as crucial for helpful hints~\cite{greifenstein.2023}. 
\autoref{fig:dialogue-example} also shows how the GenAI tutor can fail: after the child answers \say{yes}, the tutor confirms a nesting that is missing on the screen.
Nonetheless, 98.4\% (1,151/1,170) of the tutor's turns were intelligible, and only one child could not clearly understand the voice. Yet, this is a precondition for using speech to lower reading and writing demands~\cite{chen.2024a,greifenstein.2023a}. 
Only 30.4\% (7/23) of children agreed that the tutor helped them understand an error, and 53.2\% (622/1,170) of turns were judged helpful (overall impression).
Although a video had explained that responses could be incorrect, 39.1\% (9/23) of the children trusted the tutor's answers to always be correct, and only 30.4\% (7/23) reported noticing mistakes. Similarly, ninth graders rated a voice tutor relatively high on competence, although 28.6\% (119/416) of its feedback was incorrect~\cite{jacobs.2025c}.

\paragraph{Short Voiced Prompts from Young Learners.}
Children's verbalized prompts were brief: 29.2\% (355/1,216) comprised at most three words, and \say{yes} was the most frequent one (RQ2). Most prompts, 90.8\% (1,104/1,216), were responses (RES) to the tutor, and only 4.3\% (52/1,216) were standalone queries (STA). 
One reason may be the GenAI tutor, which included an open question in 54.9\% (642/1,170) of its responses or a simple closed question in 17.5\% (205/1,170) of its responses.
An additional factor could be the children's ability to put what they need into words. The six preparatory Scratch lessons were their first programming lessons, so their CS vocabulary for naming blocks or describing errors was likely limited. 
Age may also contribute, although we did not assess language skills. 
In her review of help-seeking in children, \citet{gall.1981} notes that young children often indicate verbally that they have understood when they have not. Instead, they may express their lack of understanding nonverbally. The GenAI tutor used in this study does not capture nonverbal expressions.
Therefore, some \say{yes} responses may not indicate understanding or may be misleading (as in \autoref{fig:dialogue-example}).

Children also have to learn which context a GenAI tool needs. In the text chat of Scratch Copilot, the assistant often misunderstood vague questions such as \say{Make it bigger?} about an on-screen character~\cite{druga.2025}.
Similarly, children aged 7-12 learned over time how to phrase questions to a Google Home device~\cite{butler.2024}. Screen sharing could make such references resolvable, but only if the tutor correctly perceives the program state.

\paragraph{Screen Sharing Without Program State}
The gap between pilot and classroom has several possible explanations. In the pilot, an adult described one inserted error at a time through a studio microphone (\autoref{sec:tutor-setup}). In the classroom, children answered briefly, interrupted the tutor, and the microphone captured unintended speech, conditions that impaired speech recognition in another spoken tutor for children~\cite{liu.2025b}. Because the model is a black box, we cannot determine whether its errors originated from the screen frames, the children's speech, or something else.

\section{Threats to Validity} \label{sec:threats}

Some limitations should be noted when interpreting our results.
For example, the study involved children from four classes at one school, using German and one tutor setup. Voluntary participation and space limitations may have biased the sample. Moreover, prior Scratch instruction does not establish equal individual knowledge or competency. Teacher assistance and peer conversations may also have affected the interactions.
Another aspect is that the main coder (and second author) taught the classes, conducted the pilot, and ran the classroom sessions. These overlapping roles may have influenced judgments and led to bias. The first author independently coded four sessions and reviewed the remaining assignments, and both authors resolved disagreements by consensus. However, no inter-rater agreement estimate is available.
It should further be noted that the analysis describes turns nested within sessions rather than independent observations of distinct errors, i.e., several ratings can concern one and the same unresolved error (e.g., \autoref{fig:dialogue-example}). Finally, the questionnaire only captured retrospective experiences of the lesson, including technical difficulties.

\section{Conclusions} \label{sec:conclusions}
We studied a multimodal GenAI tutor (Gemini 2.5 Flash Native Audio Preview in Google AI Studio) with screen sharing in Scratch lessons of grades 5 and 6, combining an expert pilot, a content analysis of 1,170 tutor turns and 1,216 voice prompts from 23 children in a classroom, and a questionnaire. 
Regarding RQ1, the tutor's responses were intelligible and actionable, and no single turn revealed a complete solution, but they missed existing errors in 61.5\% (371/603) of error-related turns, far more often than in the pilot, and often did not locate them. 
For RQ2, children's voice prompts were brief and mostly responded to the tutor. Children regulated the pace through pause requests and interruptions. 
Concerning RQ3, children rated the voice more favorably than the content, and many trusted answers that were not always correct.
The present study contributes (1) an evaluation scheme adapted to verbal interactions with screen-sharing data as context, (2) a classroom analysis showing that screen access did not ensure tutor responses matching the program state in our setting, and (3) insights into children's experiences and perspectives.
Future work should develop custom multimodal GenAI tutor variants with access to the project state and benchmark multimodal GenAI under authentic classroom conditions.

\bibliographystyle{ACM-Reference-Format}
\bibliography{bib}

\end{document}